\documentclass{config/egpubl}
\usepackage{config/sca2026}
\SpecialIssuePaper

\usepackage[T1]{fontenc}
\usepackage{config/dfadobe}  

\biberVersion
\BibtexOrBiblatex
\usepackage[
    backend=biber,
    bibstyle=config/EG,
    citestyle=alphabetic,
    backref=true
]{biblatex}
\electronicVersion
\PrintedOrElectronic
\ifpdf \usepackage[pdftex]{graphicx} \pdfcompresslevel=9
\else \usepackage[dvips]{graphicx} \fi
\usepackage{config/egweblnk} 

\usepackage{amssymb}
\usepackage{mathtools}
\usepackage{booktabs}
\usepackage{multirow}
\usepackage{balance}  

\makeatletter
\def\@sect#1#2#3#4#5#6[#7]#8{%
 \ifnum #2>\c@secnumdepth
  \def\@svsec{}%
 \else
  \refstepcounter{#1}
  \edef\@svsec{\csname the#1\endcsname.\hskip 0.5em}%
 \fi
 \@tempskipa #5\relax
 \ifdim \@tempskipa>\z@
  \begingroup #6\relax
   \ifnum #2=\@ne
    \ifSFB@appendix
     \ifSFB@appendixstar
      \@hangfrom{\hskip #3\relax}{\interlinepenalty \@M
       Appendix \csname the#1\endcsname\par}%
     \else
      \@hangfrom{\hskip #3\relax}{\interlinepenalty \@M
       Appendix \csname the#1\endcsname.\hskip 0.5em #8\par}%
     \fi
    \else
     \@hangfrom{\hskip #3\relax\@svsec}{\interlinepenalty \@M #8\par}%
    \fi
   \else
    \@hangfrom{\hskip #3\relax\@svsec}{\interlinepenalty \@M #8\par}%
   \fi
  \endgroup
  \csname #1mark\endcsname{#7}%
  \addcontentsline{toc}{#1}{\ifnum #2>\c@secnumdepth \else
   \protect\numberline{\csname the#1\endcsname}\fi #7}%
 \else
  \def\@svsechd{#6\hskip #3\@svsec #8
   \csname #1mark\endcsname{#7}
   \addcontentsline{toc}{#1}{\ifnum #2>\c@secnumdepth \else
    \protect\numberline{\csname the#1\endcsname}\fi#7}}%
 \fi
 \@xsect{#5}}
\makeatother

\usepackage{xcolor}

\title{Generalized Audio-Driven Synthesis of Precise Drummer Motion}

\author[Iñesta et al.] 
{
    \parbox{\textwidth}{\centering
        Álvaro G. Iñesta$^{1}$,\;\;\;
        Mattia Ryffel$^{1}$,\;\;\;
        Amit H. Bermano$^{1,2,3}$,\;\;\;
        Robert W. Sumner$^{1,3}$,\;\;\;
        Martin Guay$^{1}$
        \\
        \vspace{10pt}
        $^1$DisneyResearch|Studios, Switzerland
        \\
        $^2$The Blavatnik School of Computer Science and AI, Tel-Aviv University
        \\
        $^3$ETH Zürich, Switzerland
    }
}

\begin{document}

\teaser{
\vspace{-15pt}
 \includegraphics[trim={0 0 0 5cm},clip,width=.95\linewidth]{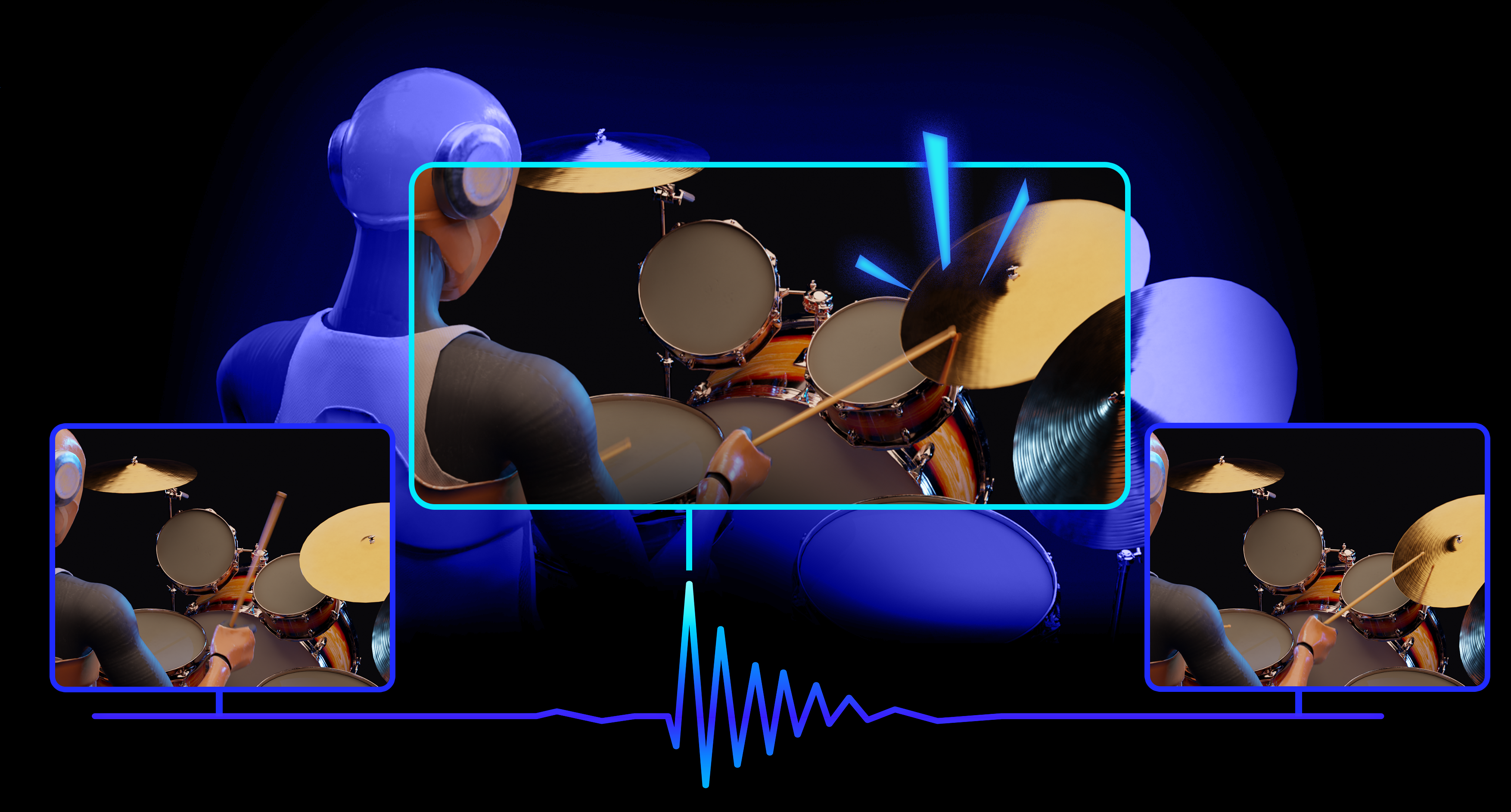}
 \centering
  \caption{\textbf{We present a precise and robust diffusion-based approach to synthesizing drumming motion from audio.}
    Given an input audio waveform (bottom), our diffusion model generates full-body motion (figure insets) with centimeter-level spatial precision and sub-frame temporal alignment. Our dual-objective loss enables systematic optimization while our performance metrics provide rigorous evaluation of both spatial and temporal quality.}
\label{fig:teaser}
}

\maketitle
\thispagestyle{empty}
\makeatletter
\renewcommand{\@oddfoot}{\hfil\thepage\hfil}
\renewcommand{\@evenfoot}{\hfil\thepage\hfil}
\makeatother

\addtolength{\topmargin}{-0.6cm} 
\addtolength{\headsep}{0.5cm}
\addtolength{\footskip}{0.5cm}
\addtolength{\textheight}{2.0cm}


\begin{abstract}
Music-driven character animation enables and enhances transformative applications in entertainment and interactive education. However, synthesizing realistic drumming motion from audio remains challenging due to the inherent tension between high-acceleration dynamics and the need for extreme spatial-temporal precision. Existing approaches, often reliant on motion matching or MIDI input, struggle with generalizing to diverse real-world audio.
Moreover, the field lacks standardized evaluation metrics capable of distinguishing precise drumming from noisy motion.
In this paper, we introduce a generative diffusion framework featuring a dual-objective loss function that decouples skeletal integrity from drumstick precision, thus enabling centimeter-level stick precision without sacrificing natural body dynamics. Additionally, leveraging our own dataset and data augmentation strategy, the model generalizes to non-curated, in-the-wild audio.
To rigorously evaluate performance, we propose two novel metrics: an impact-to-target distance to quantify spatial precision and an audio-motion correlation score to assess temporal alignment.
Our quantitative analysis and user studies demonstrate that our system generates high-quality motion that is often indistinguishable from ground-truth performances.
\\



\end{abstract}

\section{Introduction}\label{sec.intro}
Synthesizing musical performances traditionally relies on resource-intensive methods, such as manual animation and motion capture, which lack the scalability needed for large-scale applications.
Data-driven generative models offer a promising alternative and have shown a great success in automating high-quality motion synthesis~\cite{Tevet2022,Tseng2023,Li2025,Qiu2025}.
However, synthesizing realistic instrumental performances directly from audio poses additional challenges beyond general motion generation. Unlike speech-driven gestures~\cite{Ng2024,Kim2024} or music-conditioned dance~\cite{Siyao2022,Tseng2023,Wang2025}, instrument-driven motion demands extreme spatial and temporal precision: the motion of the performer must be perfectly aligned with the instrument's physical layout and with the sound events.
The specific challenges of this domain vary significantly by instrument class. For example, piano~\cite{Wang2024} and guitar~\cite{PerezCarrillo2019,Xu2024,Canales2025} require precise fine-grained finger motion, while cello~\cite{Qiu2025} additionally demands alignment between the bow and the instrument.
Percussive instruments, and drums in particular, pose a fundamentally unique set of challenges:
\begin{enumerate}
    \item \textbf{Acoustic ambiguity in percussive onsets.}
    Unlike melodic instruments, drum audio consists of noisy bursts. The similarity between the sounds of the different components (e.g., a hi-hat and a crash) makes it difficult to extract discriminative features for driving motion, hindering generalization across different drum kits and room acoustics.
    \item \textbf{High accelerations.}
    Drumming involves precise and high velocity motion, with stick velocities often exceeding 10~m/s~\cite{Dahl2004}. Conventional motion synthesis results in dispersion jitter at the limb extremities. It is a challenge to remove the jitter while preserving the high-speed dynamics, without smoothing out the motion altogether.
    \item \textbf{Arm ambiguity.} In drumming, the same acoustic event may be played by either hand. This creates a one-to-many problem where the model must not only synthesize precise motion, but also make high-level coordination decisions.
\end{enumerate}

Due to these complexities, existing generative frameworks remain ill-equipped to bridge the gap between noisy percussive signals and high-velocity physical execution. While current audio-driven systems excel at capturing the fluid, low-frequency trajectories of melodic instruments~\cite{Tseng2023,Qiu2025}, they lack the capability to resolve acoustic ambiguities in drum onsets or to maintain stability during extreme accelerations. Consequently, there remains a critical need for a synthesis approach that can decouple high-level coordination from low-level physical precision while attending to inherently noisy audio signals.

We present an audio-driven generative motion model that achieves high-quality drumming motion at scale, solving the aforementioned challenges.
We describe the steps required to establish this first successful baseline and introduce three core contributions:
\begin{itemize}
    \item \textbf{High-fidelity dataset and augmentation.}
    We curate a motion capture dataset with 3.5+ hours of professional drumming at a high frame rate, covering a diverse range of styles, tempos, and intensities.
    To bridge the gap between studio recordings and in-the-wild audio, we introduce a domain-specific data augmentation strategy that ensures generalization across different acoustics.
    \item \textbf{New loss for high-frequency precision.}
    Since drumstick-tip trajectories are the primary drivers of acoustic events, they demand a higher level of precision than general skeletal motion.
    To address this, we introduce a dual-objective loss that decouples body motion from stick-tip trajectories: body joints are parameterized by rotations to ensure skeletal integrity, while stick tips are directly supervised in Cartesian coordinates for centimeter-level precision.
    \item \textbf{New metric for percussive motion.}
    Previous metrics to measure the quality of audio-driven motion fail to distinguish between precise hits and noisy jitter across datasets, impairing principled evaluation.
    Our novel Percussive Alignment Score (PAS), quantifies audio-motion onset correspondence using a parameterized Gaussian kernel, exhibiting near-zero values for random motion and approaching unity for precise performance.
    In addition, as drumming requires hitting accurate points on the instruments, we introduce an Impact Point Deviation (IPD) score, explicitly quantifying stick-tip hitting fidelity.
\end{itemize}

Extensive evaluations demonstrate that our approach generates highly precise motion, with human perceptual studies confirming that our synthesized motion is nearly indistinguishable from ground truth.
Furthermore, we prove that our data augmentation strategy ensures robustness when generalizing to unseen, non-curated audio.
Lastly, we demonstrate the broader utility of our framework by repurposing the audio-to-motion pipeline as a drum transcriber.
Even without task-specific training, this motion-informed approach achieves competitive results, outperforming specialized state-of-the-art transcribers on our evaluation set.

\begin{figure*}[ht]
  \centering
  \includegraphics[width=0.9\linewidth]{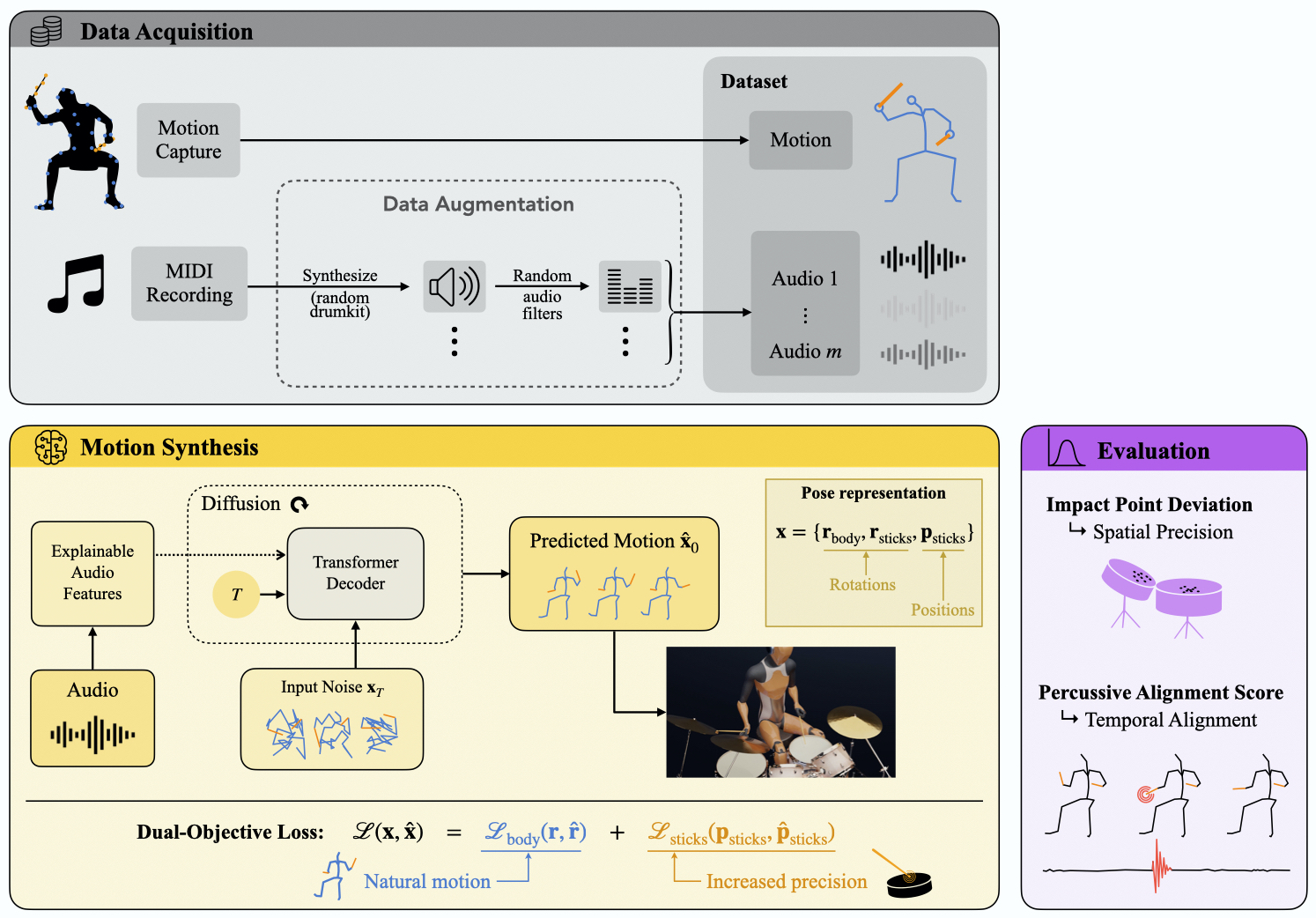}
  \caption{
    \textbf{Overview.}
    We augment the audio in our dataset resulting in motion sequences aligned with multiple audio variations to ensure generalization.
    Raw audio is then processed into explainable features to condition a diffusion-based Transformer Decoder.
    We employ a hybrid pose representation, using joint rotations for the body and Cartesian positions for drumsticks.
    The model is trained via a dual-objective loss to balance natural body dynamics with high-precision stick impacts.
    Finally, performance is evaluated using Impact Point Deviations and Percussive Alignment Scores to assess spatial and temporal fidelity.
  }
  \label{fig.system}
\end{figure*}


\section{Related Work}

\paragraph*{Motion synthesis: from blending to generation.}
Traditional approaches to character animation have evolved from pose blending and state machines~\cite{Rose1998Verbs}, through motion graphs~\cite{Kovar2002}, to motion matching~\cite{Clavet2016} -- now the industry standard for game engines.
However, these methods require large databases of captured or manually authored data, and they often struggle to generalize to highly expressive or specialized tasks.
To overcome these limitations, recent research has shifted towards data-centric methods.
In particular, generative motion models learn distributions rather than deterministic mappings, leading to more expressive and diverse results.
Recently, diffusion models~\cite{SohlDickstein2015,Ho2020} have emerged as the leading architecture for high-fidelity synthesis, offering superior diversity than GANs~\cite{Goodfellow2014} or VAEs~\cite{Kingma2013}.

\paragraph*{Audio-conditioned motion synthesis.}
Conditioning motion generation on audio has proven effective across multiple domains.
In particular, diffusion has successfully synthesized fluid motion for dance~\cite{Siyao2022,Tseng2023,Wang2025} and melodic instruments like the cello~\cite{Qiu2025}.
However, these frameworks do not translate directly to percussion: unlike continuous melodies, percussive audio consists of discrete bursts, and the resulting motion involves high-acceleration trajectories rather than the temporal smoothness typically produced by existing models.

\paragraph*{Drumming-specific approaches.}
Percussive instruments pose fundamentally different challenges than melodic instruments: noisy and discrete acoustic bursts, high accelerations requiring centimeter-level precision, and one-to-many mappings between audio and arm assignments.
The combination of spatial precision (hitting the correct drum component), temporal precision (at exactly the right moment), and natural full-body motion represents a unique synthesis challenge. To our knowledge, only two data-centric approaches have addressed drumming motion synthesis.
Shahid et al.~\cite{Shahid2025} trained a humanoid robot to play drums using reinforcement learning with MIDI input.
While rhythmically accurate, the resulting motion lacks human-like expressivity due to the mechanical constraints of the robot.
Kyriakou et al.~\cite{Kyriakou2025} employed a hybrid system combining LSTM-driven arms, motion-matching for torso and face, and deterministic functions for legs, also using MIDI input.
Despite the absence of mechanical constraints, the reliance on a database search and heuristics still limited the diversity and physical plausibility of the generated motion.

In this paper, we propose an end-to-end diffusion-based approach to map raw audio to drumming motion, and our approach differs from previous in several key points. First, we use a novel loss formulation designed for the dual requirements of skeletal plausibility and end-effector precision. The use of audio as input rather than MIDI generalizes to more music applications. Most songs are either unavailable in MIDI (with automatic audio-to-MIDI transcription for drums remaining unreliable~\cite{Zehren2023}), or are misaligned temporally with the audio as they were manually authored.
Combined with a data augmentation strategy, we achieve generalization to out-of-distribution music, including non-curated recordings retrieved from various sources. Lastly, we introduce new evaluation metrics that enable systematic comparison.


\section{Method}

We begin by defining our motion representation and instrument configuration (Section~\ref{sec.instrument-motion}), and then detailing our data collection and augmentation strategy (Section~\ref{sec.dataset}). We explain our audio feature extraction approach (Section~\ref{sec.audio}) before introducing our diffusion model architecture and our dual-objective loss function (Sections~\ref{sec.architecture} and~\ref{sec.loss}). Finally, we provide training details and inference procedures for generating long-form performances in Section~\ref{sec.training}.
Figure~\ref{fig.system} shows an overview of our method.

\subsection{Instrument Configuration and Motion Representation}\label{sec.instrument-motion}

\begin{figure}[t!]
  \centering
  \includegraphics[width=\linewidth]{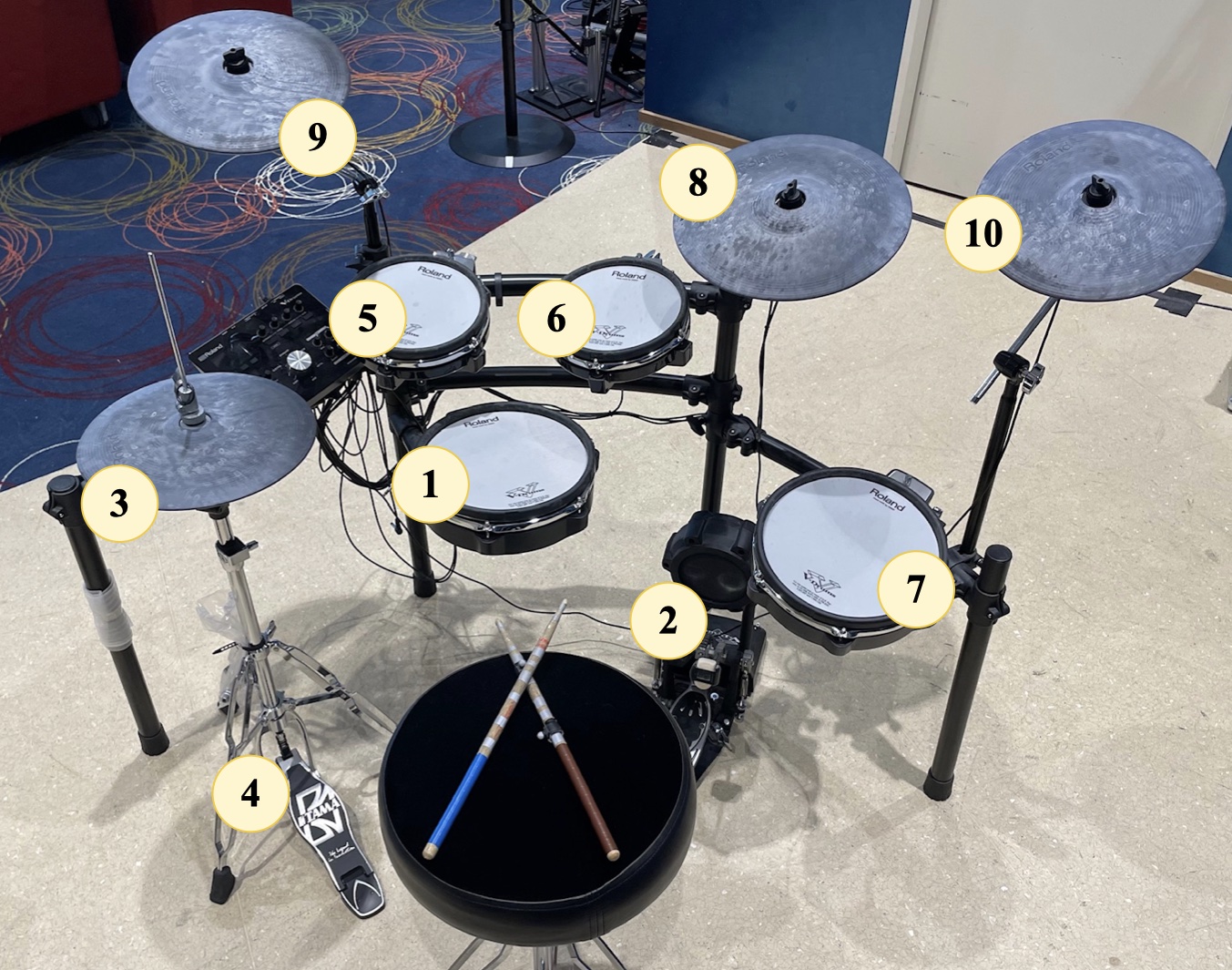}
  \caption{
    \textbf{Standard 10-component drum kit used in our data collection.}
    (1) Snare, (2) bass/kick drum, (3) hi-hat, (4) hi-hat pedal, (5-6) high toms, (7) floor tom, (8) ride cymbal, (9-10) crash cymbals.
  }
  \label{fig.drumkit}
\end{figure}

We use a standard 10-component drum kit comprised of snare, bass drum (kick), hi-hat, hi-hat pedal, two high toms, floor tom, ride cymbal, and two crash cymbals (Figure~\ref{fig.drumkit}). This common setup is used across multiple genres from jazz to rock, providing sufficient coverage of popular drumming sound patterns while remaining computationally tractable.

We represent drumming motion as sequences of $n$ poses sampled at frame rate $f = 120$~Hz. The high frame rate is critical for capturing the high-velocity motion, since stick velocities can exceed 10 m/s~\cite{Dahl2004}.
Each of the $n$ poses represents the drummer's body and drum sticks. The body is comprised of 27 joints modeled using the continuous 6D rotation representation~\cite{Zhou2019}: $\mathbf{r}_\mathrm{body} \in \mathbb{R}^{n\times27\times6}$.
We assume a fixed hip position as the drummer remains seated throughout the performance.

We model the drum sticks as rigid bodies connected to a hand, via orientations $\mathbf{r}_\mathrm{sticks} \in \mathbb{R}^{n\times2\times6}$.
Furthermore, we found that adding the tip of the sticks as additional coordinates to infer, $\mathbf{p}_\mathrm{sticks} \in \mathbb{R}^{n\times2\times3}$, significantly improved the accuracy of the hits.
We put special attention to the tip because this is often the part of the stick that impacts the drums (some components, such as the cymbals, are occasionally hit with other parts of the stick, while the pedals are always triggered by the feet).
Note that we normalize $\mathbf{p}_\mathrm{sticks}$ over our whole dataset and therefore remains dimensionless and in the same order of magnitude as the rotations.

A complete motion sequence is thus represented as $\mathbf{x} = \{\mathbf{r}, \mathbf{p}_\mathrm{sticks}\} \in \mathbb{R}^{n\times180}$, where $\mathbf{r} = \{\mathbf{r}_\mathrm{body}, \mathbf{r}_\mathrm{sticks}\}$.
Having defined our motion representation, we now describe the data collection process.

\subsection{Dataset Collection and Augmentation}\label{sec.dataset}

Existing publicly available drumming datasets are insufficient for training generative models. The largest prior dataset~\cite{Kyriakou2025a} contains approximately 35 minutes of drumming, with reported issues including ``high noise and frequent foot sliding'' due to noisy inertial sensors~\cite{Kyriakou2025}. We therefore captured our own high-quality dataset using optical motion capture.

\paragraph*{Capture setup.}
We recorded a professional jazz drummer performing on a Roland TD-25KV electronic drum kit over two days. Motion was captured using nine OptiTrack~\cite{OptiTrackMotive} cameras at 120~Hz, strategically positioned to minimize occlusions from the drum kit. The drummer wore a motion capture suit with additional markers attached to drumsticks, tracked as rigid bodies. The drum kit was fixed to the ground to prevent drift during repeated impacts. Audio was recorded at 44.1~kHz, also in MIDI format.
MIDI is used only for evaluation, not for training, as we design our model to work with audio input for real-world deployment (see `Data augmentation' paragraph below).

\paragraph*{Dataset composition.}
Our final cleaned dataset contains 3 h 30 min 53 s (1,518,450 frames) of aligned audio and motion.
Our recordings span three categories designed to ensure coverage of fundamental techniques, common patterns, and musical diversity:
\begin{itemize}
  \item \textbf{Fundamentals}: individual drum components played at various tempos and intensities (1-3 minutes each), plus all combinations of two and three components (6 minutes in total), totaling approximately 30 min.
  \item \textbf{Grooves}: 17 common patterns and styles (rock, disco, funk, jazz, reggae, metal, etc.), each performed at three different tempos with gain variations, totaling approximately 1 h 53 min.
  \item \textbf{Songs}: improvised performances over diverse backing tracks spanning show tunes, pop, big band, swing, jazz, rock, blues, soul, K-pop, Latin pop, and country, totaling approximately \mbox{1 h 7 min}.
\end{itemize}

We deliberately excluded cross-armed playing based on our professional drummer's advice that this technique is considered inefficient and is rarely taught in modern instruction. This choice simplifies the arm ambiguity problem.

\paragraph*{Data augmentation.}\label{par.dataaug}
To ensure generalization across drum kits and acoustic environments, we apply a two-stage augmentation strategy. For each audio-motion sample, we generate $m=50$ augmented audio variants by
(1) synthesizing audio from the MIDI signal using a randomly selected acoustic or electronic drum kit, and then (2) applying 13 randomized audio effects including reverb, noise, equalization, compression, and pitch shift, each with strength sampled uniformly from $[0, 1]$ where 0 means the effect is not applied and 1 means maximum effect.
The drum kit and the audio effects are chosen independently at random for each sample.
During training, one of these $m$ variants of the dataset is randomly selected per epoch, expanding our acoustic diversity without additional motion capture. Section~\ref{subsec.wild} demonstrates this strategy is critical for robust generalization to out-of-distribution audio.

With paired audio-motion data collected, we next describe how we extract meaningful features from the audio signal to condition motion generation.

\subsection{Audio Feature Extraction}\label{sec.audio}

Our model processes raw audio through a custom feature extractor producing 44-dimensional feature vectors per frame, synchronized with the motion frame rate of $f=120$~Hz. Unlike approaches using pretrained audio encoders like Jukebox~\cite{Dhariwal2020}, we employ interpretable handcrafted features specifically tailored to percussive audio:

\begin{itemize}
  \item \textbf{Onset detection} (binary): indicates whether a drum component was hit at a given frame.
  \item \textbf{Beat tracking} (binary): indicates the beat, providing rhythmic context.
  \item \textbf{Amplitude envelope} (continuous): captures the audio intensity.
  \item \textbf{Spectral centroid} (continuous): average frequency indicating brightness, useful for distinguishing cymbals from drums.
  \item \textbf{MFCCs} (40-dimensional, continuous): Mel-frequency spectral coefficients capturing timbral characteristics that distinguish individual drum components.
\end{itemize}

Note that the input audio to the feature extractor must be a drums-only track.
For deployment on polyphonic audio containing multiple instruments, we first extract the drum stem using state-of-the-art source separation tools, such as Demucs~\cite{Defossez2019} or Spleeter~\cite{Hennequin2020}, before applying our feature extractor.
Further details on audio feature extraction can be found in Appendix~\ref{app.audio-feats}. These audio features serve as the conditioning signal for our generative model, which we describe next.

\subsection{Diffusion Model Architecture}\label{sec.architecture}

We adapt the EDGE architecture~\cite{Tseng2023}, a diffusion probabilistic model originally designed for audio-conditioned dance generation. Diffusion models have emerged as state-of-the-art for music-driven motion synthesis due to their ability to capture multi-modal distributions and generate diverse, high-quality motion that captures the stochastic nuances of natural performance. Unlike deterministic regressors that produce averaged motion or motion matching systems constrained by database coverage, diffusion models learn the full distribution of plausible motions conditioned on audio.

Let $\mathbf{x}_0 = \{ \mathbf{r}, \mathbf{p}_\mathrm{sticks} \}\in \mathbb{R}^{n \times 180}$ denote ground-truth motion and $\mathbf{c} \in \mathbb{R}^{n \times 44}$ the extracted audio features.
The forward diffusion process progressively adds Gaussian noise over $T$ timesteps similar to Ho et al.~\cite{Ho2020}.
The reverse process employs a Transformer-based network $\epsilon_\theta(\mathbf{x}_t, t, \mathbf{c})$ that predicts the noise to be removed at each step, with audio conditioning integrated via cross-attention mechanisms. We refer to Tseng et al.~\cite{Tseng2023} for detailed architectural specifications (note that their architecture uses Jukebox~\cite{Dhariwal2020} for feature extraction, while we use our custom extractor).
At inference, we sample $\mathbf{x}_T \sim \mathcal{N}(0, \mathbf{I})$ from pure noise and progressively denoise for $T=1000$ steps. Using the DDIM sampler~\cite{Song2020} to skip steps, we only require about 5 steps.

While the EDGE architecture provides a strong foundation for audio-conditioned motion synthesis, standard diffusion training objectives do not adequately address the unique precision requirements of drumming. We therefore introduce a novel loss formulation.

\subsection{Dual-Objective Loss Function}\label{sec.loss}

Motion diffusion models often supervise all joints uniformly as rotations, using forward kinematics (FK) for positions. This ensures skeletal consistency but struggles with end-effector precision, as rotation errors propagate through kinematic chains.

For drumming, stick-tip precision is paramount: a few centimeters error determines hitting the drum versus missing. Conversely, body motion benefits from rotational parameterization for fixed bone lengths and natural posing. We introduce a \textbf{dual-objective loss} decoupling these requirements:

\begin{equation}\label{eq.dual_loss}
\mathcal{L}(\mathbf{x}_0, \hat{\mathbf{x}}_0) =
\lambda_r \| \mathbf{r} - \hat{\mathbf{r}} \|_2^2 +
\lambda_p \| \mathbf{p}_\mathrm{sticks} - \hat{\mathbf{p}}_\mathrm{sticks} \|_2^2
\end{equation}
where $\mathbf{x}_0 = \{ \mathbf{r}, \mathbf{p}_\mathrm{sticks} \}$ and $\hat{\mathbf{x}}_0 = \{ \hat{\mathbf{r}}, \hat{\mathbf{p}}_\mathrm{sticks} \}$ correspond to ground-truth and predicted motion, respectively. We use the loss weights $\lambda_r = 0.5$ and $\lambda_p = 1$, which we found to provide the best results.
Through this loss, the model directly optimizes stick-tip positions in Cartesian space while maintaining skeletal constraints on the body. Section~\ref{subsec.hqmotion} demonstrates this significantly outperforms rotation-only baselines.
While the original EDGE architecture~\cite{Tseng2023} employs velocity, forward-kinematics, and foot-contact losses, we found that these additional terms did not provide a noticeable improvement in drumming motion quality and we therefore omit them.

\subsection{Training Details}\label{sec.training}

We train on sequences of $n=120$ frames (1 second at 120~Hz). This window size balances temporal context with generation quality: shorter sequences fail to capture rhythmic patterns and produce noisy output, while longer sequences lead to over-smoothing and loss of high-frequency detail in the high-acceleration stick motion.
Our dataset then consists of 25,000+ motion sequences (with 0.5 s overlap between consecutive sequences) and 50 audio variants per sequence.
We use approximately 3,000 sequences for test (these sequences correspond to different takes and therefore have no overlap with any training sequences).    
We employ the Adam optimizer with learning rate $3\cdot10^{-4}$ and batch size 128, training for 6000 epochs on an NVIDIA GeForce RTX 3090 (approximately 48 h).

For \textbf{long-form motion synthesis} at inference time, we employ a sliding window approach with 50\% overlap between consecutive windows. Adjacent windows are blended using spherical linear interpolation for rotations and linear interpolation for positions, following the stitching procedure from Tseng et al.~\cite{Tseng2023}. This enables generation of arbitrary-length performances while maintaining temporal coherence.

Having described our complete synthesis pipeline, we now turn to the critical question of evaluation: how can we rigorously quantify the quality of generated drumming motion, particularly the temporal and spatial precision that distinguish convincing from unconvincing performance?


\section{Evaluation Metrics}

A fundamental challenge in evaluating drumming motion is measuring both spatial precision (does the drummer hit the correct drum component?) and temporal alignment (does the motion synchronize with the audio?). Existing metrics for dance or general motion do not capture either aspect adequately. We introduce two complementary metrics designed for precision evaluation.

\subsection{Spatial Precision: Impact Point Deviations}

We define an \emph{impact point} as the position of the tip of the stick when it hits a drum.
To measure the spatial precision of drumming, we evaluate the deviation between the impact points of generated motion and ground truth.
To do this, we first extract the position of the sticks at every audio onset, and then use density-based clustering to remove faulty detections (see Appendix~\ref{app.IPD} for further details).
The mean impact point per drum component is then calculated and compared to ground truth via Euclidean distance.

\subsection{Temporal Precision: Percussive Alignment Score (PAS)}\label{sec.bas}

In the case of drumming, the most relevant indicator of temporal alignment is the correspondence between audio and motion onsets: the times at which we \emph{hear} and \emph{see} a drum impact.
Let $\mathcal{A}$ denote audio onset times and $\mathcal{M}$ denote motion onset times.
Audio onsets can be computed from the audio signal, while motion onsets are peaks in the linear acceleration of the drum sticks. In Appendix~\ref{app.DAS}, we provide further details on how to compute $\mathcal{A}$ and $\mathcal{M}$.

Prior work measured temporal alignment with the delay between audio onsets and the closest motion onset~\cite{Kyriakou2025}.
However, delays audio and motion are normally recorded on separate devices, and therefore data recorded with different setups would have different delays.
To solve this issue, we define the Percussive Alignment Score (PAS), similar to the metric introduced in Siyao et al.~\cite{Siyao2022}:

\begin{equation}\label{eq.bas}
\mathrm{PAS}(\mathcal{A}, \mathcal{M}) \coloneqq
\frac{1}{|\mathcal{A}|}
\sum_{s \in \mathcal{A}} \exp \left\{ - \left( \frac{
\min_{r \in \mathcal{M}} | r - s |}{\alpha} \right)^\beta \right\}
\end{equation}

where $\alpha=40$~ms, $\beta=7$. This function computes soft correspondence scores using a generalized Gaussian kernel.
The choice of values for $\alpha$ and $\beta$ ensure that small audio-motion delays score high, while perceptually asynchronous delays score low (see Appendix~\ref{app.DAS} for a detailed discussion).
Note that $\mathrm{PAS} \in [0,1]$.
High-quality data will achieve values close to 1, while noisy data with bad audio-motion alignment will score low (see examples in Appendix~\ref{app.DAS}).

Together, the impact point deviations and the PAS provide a comprehensive evaluation framework for drumming motion.
Next, we present our results and use these metrics to validate them.


\section{Results}

We first validate our contributions with different metrics and comparisons.
In Section~\ref{subsec.hqmotion}, we focus on assessing the quality of the generated motions. 
Then, in Section~\ref{subsec.wild}, we test generalization with in-the-wild audio.
Lastly, in Section~\ref{subsec.transcriber}, we discuss an application of our model to music analysis: transcription from audio to MIDI.

\begin{figure}[t!]
  \centering
  \includegraphics[width=\linewidth]{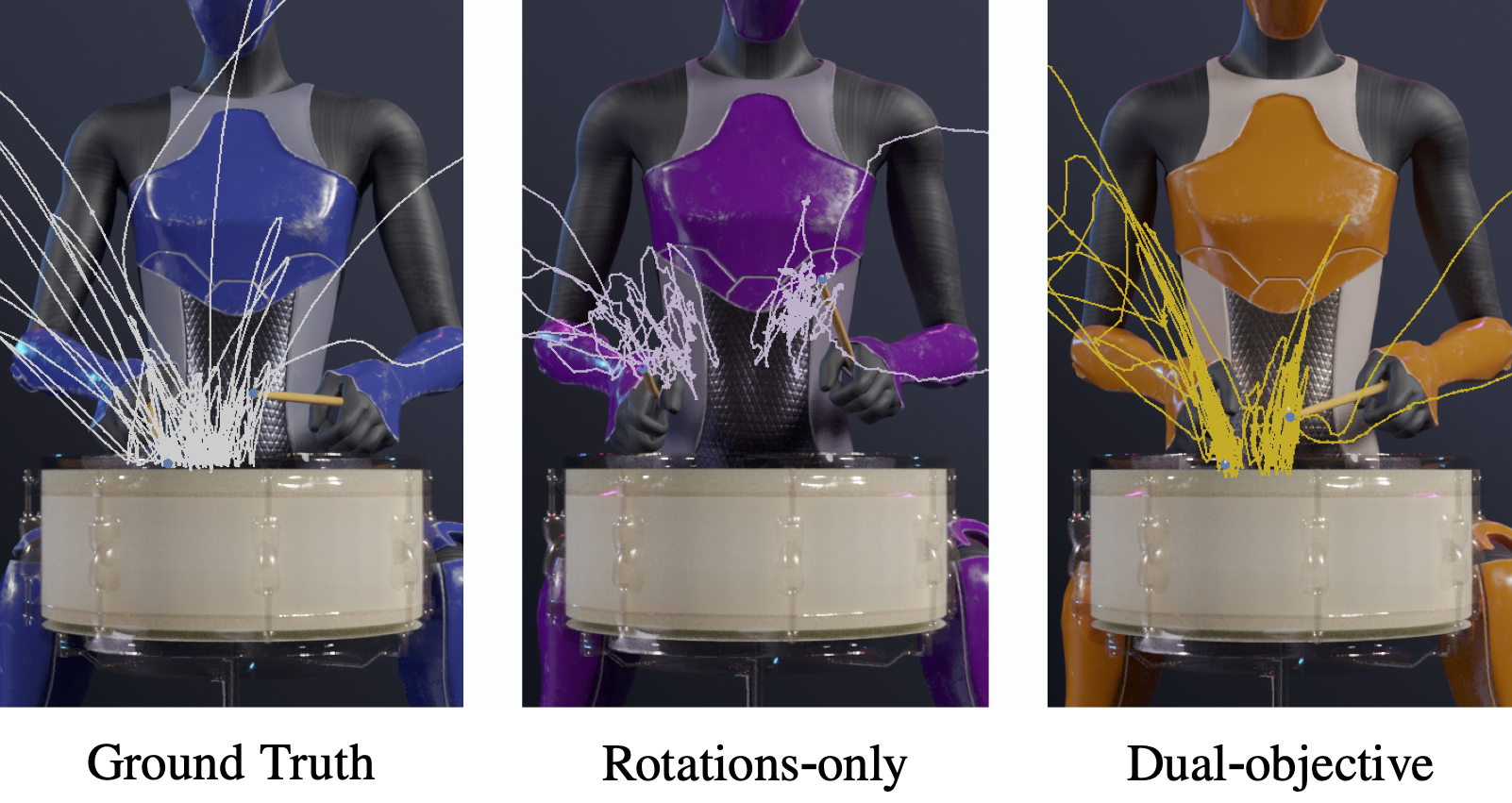}
  \caption{
    \textbf{Dual-objective loss achieves superior spatial precision.}
    Stick-tip trajectories during a snare roll. Ground truth (left), rotations-only model (center), and our dual-objective model (right).
    Tight clustering on the drum surface demonstrates consistent impact points.
  }
  \label{fig.motion_paths_rolls}
\end{figure}

\begin{table}[t!]
\caption{Impact point deviation per drum component. Values are Euclidean distances (cm) between the impact point centroids of our generated motion and ground-truth over the test set. Lower values are indicative of higher-quality motion. All standard errors are below 0.01 cm for the rotations-only model and below 0.006 cm for the dual-objective model (ours).}
\label{tab.spatial}
\centering
\begin{tabular}{lccc}
\toprule
Component & Rot. Only (cm) $\downarrow$ & Ours (cm) $\downarrow$ \\
\midrule
Snare & 14.5 & 0.5 \\
High tom (L) & 5.0 & 1.2 \\
High tom (R) & 8.2 & 3.0 \\
Floor tom & 6.4 & 3.3 \\
\midrule
\textbf{Mean (drums)} & 8.5 & \textbf{2.0} \\
\midrule
Hi-hat & 2.5 & 1.3 \\
Ride cymbal & 12.7 & 0.8 \\
Crash cymbal (L) & 15.2 & 2.2 \\
Crash cymbal (R) & 2.7 & 3.1 \\
\midrule
\textbf{Mean (cymbals)} & 8.3 & \textbf{1.8} \\
\bottomrule
\end{tabular}
\end{table}


\subsection{Motion Quality}\label{subsec.hqmotion}
We use impact point deviations, PAS, and user preference scores to compare ($i$) ground truth, ($ii$) a rotations-only model (all joints modeled as rotations, with stick-tips calculated via FK), and ($iii$) our dual-objective model (body rotations + stick positions). Both models use identical architectures, data, and hyperparameters (see Section~\ref{sec.training}).

\paragraph*{Spatial Precision.}
Figure~\ref{fig.motion_paths_rolls}, shows the motion paths of the stick tips during a snare roll (the full animation is included in the Supplementary Video).
The impact points of the rotations-only model are far from the drum surface and drift over time.
In contrast, our dual-objective model is able to consistently hit the same spot, achieving visually similar results to ground truth.
Table~\ref{tab.spatial} contains the impact point deviations between ground truth and generated motion for each drum component.
The average deviation is 8.4 cm for the rotations-only model, but only 1.9 cm for our dual-objective model.
For reference, the diameter of each drum component is around 25 cm, while each stick tip may travel up to 8 cm in between consecutive frames (top velocities $\sim 10$ m/s \cite{Dahl2004} recorded at 120 fps).
In Appendix~\ref{app.IPD}, we show the overlap between clusters of ground-truth and generated impact points.

\begin{figure}[t!]
  \centering
  \includegraphics[width=\linewidth]{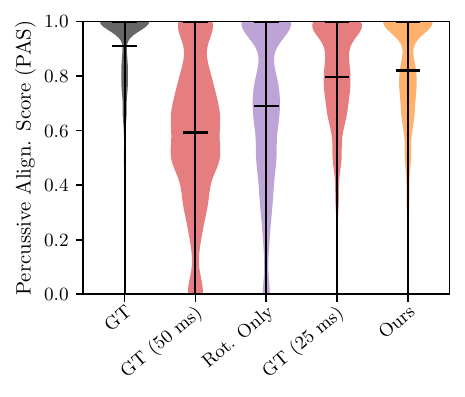}
  \caption{
    \textbf{Our model achieves temporal alignment comparable to ground-truth data.}
    PAS distributions as violin plots for GT (leftmost, black), noisy GT (second and fourth, red), rotations-only model (middle, purple), and our model (rightmost, orange), when applied to our test dataset.
    Each sample is one-second long.
    Noisy GT is obtained by adding Gaussian noise with standard deviation $\sigma$=50 ms and $\sigma=25$ ms to every motion onset time.
    When noise is added to the data, PAS decreases, indicating that the metric successfully discriminates audio-motion alignment quality.
  }
  \label{fig.das_test}
\end{figure}

\paragraph*{Temporal Alignment.}
Figure~\ref{fig.das_test} shows the PAS distribution of ground truth (GT) and the different models on our test dataset.
We also include noisy versions of GT, where independent Gaussian noise with standard deviation $\sigma$ is added to each motion onset.
First, we observe that GT (black violin) achieves scores close to unity, with an average of 0.91.
The reason for not scoring exactly 1 is inherent noise in the motion-capture data and the peak-finding algorithms used to compute $\mathcal{A}$ and $\mathcal{M}$.
Imprecise audio-motion alignment severely penalizes PAS: $\sigma=50$ ms and $\sigma=25$ ms (red violins) drop the average PAS to 0.59 and 0.80, respectively.
Notably, our model (orange violin), with an average PAS of 0.82, outperforms the rotations-only baseline (0.68) and achieves scores closer to GT than even the version with only 25 ms of Gaussian noise.

\paragraph*{User preference.}\label{sec.user_study}
To evaluate the perceptual quality of our generated motions, we conducted a two-alternative forced choice (2AFC) study. 22 users were presented with 15 pairs of 10 s motion clips with audio, and asked to select the motion they preferred (based on realism, precision, etc.).
We evaluated three models: ground truth (GT), rotations-only (RO), and dual-objective (Ours).
All data employed in the user study belongs to our test set.
Moreover, all participants passed an attention check video at the midpoint of the experiment.

\begin{figure}[t!]
  \centering
  \includegraphics[width=\linewidth]{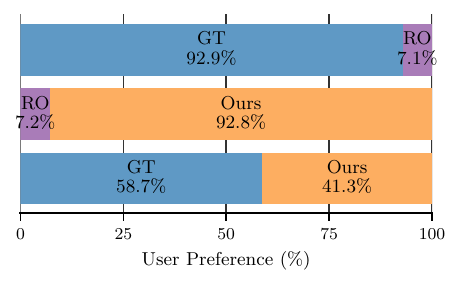}
  \caption{
    \textbf{Users show comparable preference for motion-capture data and for motion generated by our model.}
    Our generated motions (orange) strongly outperform the rotations-only model (RO, purple) with a 92.8\% preference rate ($p$-value $<0.001$).
    When compared to ground truth (GT, blue), our method achieves a 41.3\% preference rate, demonstrating no statistically significant difference from a random guess ($p$-value $0.08$).
  }
  \label{fig.user_study}
\end{figure}

The results are summarized in Figure~\ref{fig.user_study}.
As expected, users easily distinguished GT from the RO model (92.9\% preference for GT, with $p$-value $<0.001$).
Our method demonstrated a significant leap in synthesized motion quality, being preferred over RO in 92.8\% of comparisons ($p$-value $<0.001$).
Most notably, when our method was compared directly against GT, users preferred GT only 58.7\% of the time. A binomial test reveals this preference is not statistically significant ($p$-value $0.08$), suggesting that our generated motion achieves a level of realism that is perceptually comparable to captured data.

The previous quantitative analysis and the user study demonstrate that our system can achieve human-like drumming motion quality.
Next, we investigate how our data augmentation strategy enables the model to generalize to unseen data.


\subsection{In-the-Wild Audio}\label{subsec.wild}
By generating multiple variations of our captured data and training the model on such variations, we are able to apply our model to unseen audio.

\begin{figure}[t!]
  \centering
  \includegraphics[]{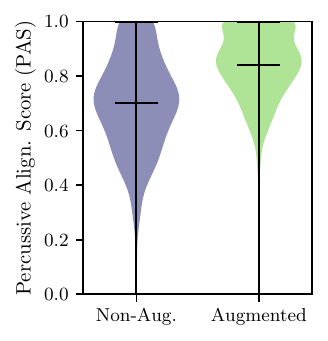}
  \caption{
    \textbf{Data augmentation ensures synthesis of high-quality motion from wild audio.}
    PAS distributions as violin plots for motion generated from 100 random audio samples from the Groove dataset. We compare our dual-objective model trained without (left) and with (right) data augmentation.
  }
  \label{fig.das_augmented}
\end{figure}

\paragraph*{Groove Dataset.}
The Groove dataset~\cite{Gillick2019} contains \mbox{13.6 hours} of drum recordings as paired audio and MIDI.
The recordings are by different performers on different-sounding drum kits, and they were recorded in a controlled environment and curated into a high-quality dataset.
As a first generalization test, we synthesize motion from 100 audio tracks from Groove, chosen at random.
We use our model trained with and without data augmentation, and we find that the non-augmented version produces very inaccurate motion, while the augmented model consistently yields a visually accurate performance, as shown in the Supplementary Video.
In particular, this difference in quality becomes clear when looking at the PAS (Figure~\ref{fig.das_augmented}):
the augmented model scores consistently higher than its non-augmented counterpart, increasing the average score from 0.7 to 0.84.

\paragraph*{Non-curated recordings.}
We further evaluated our augmentation strategy using non-curated tracks from diverse open sources. Qualitative examples demonstrating clear quality improvements are provided in the Supplementary Video.
Due to the lack of a standardized dataset and the high variance among these tracks, aggregated quantitative metrics would be unreliable and are therefore excluded from this paper.

\begin{figure}[t!]
  \centering
  \includegraphics[width=\linewidth]{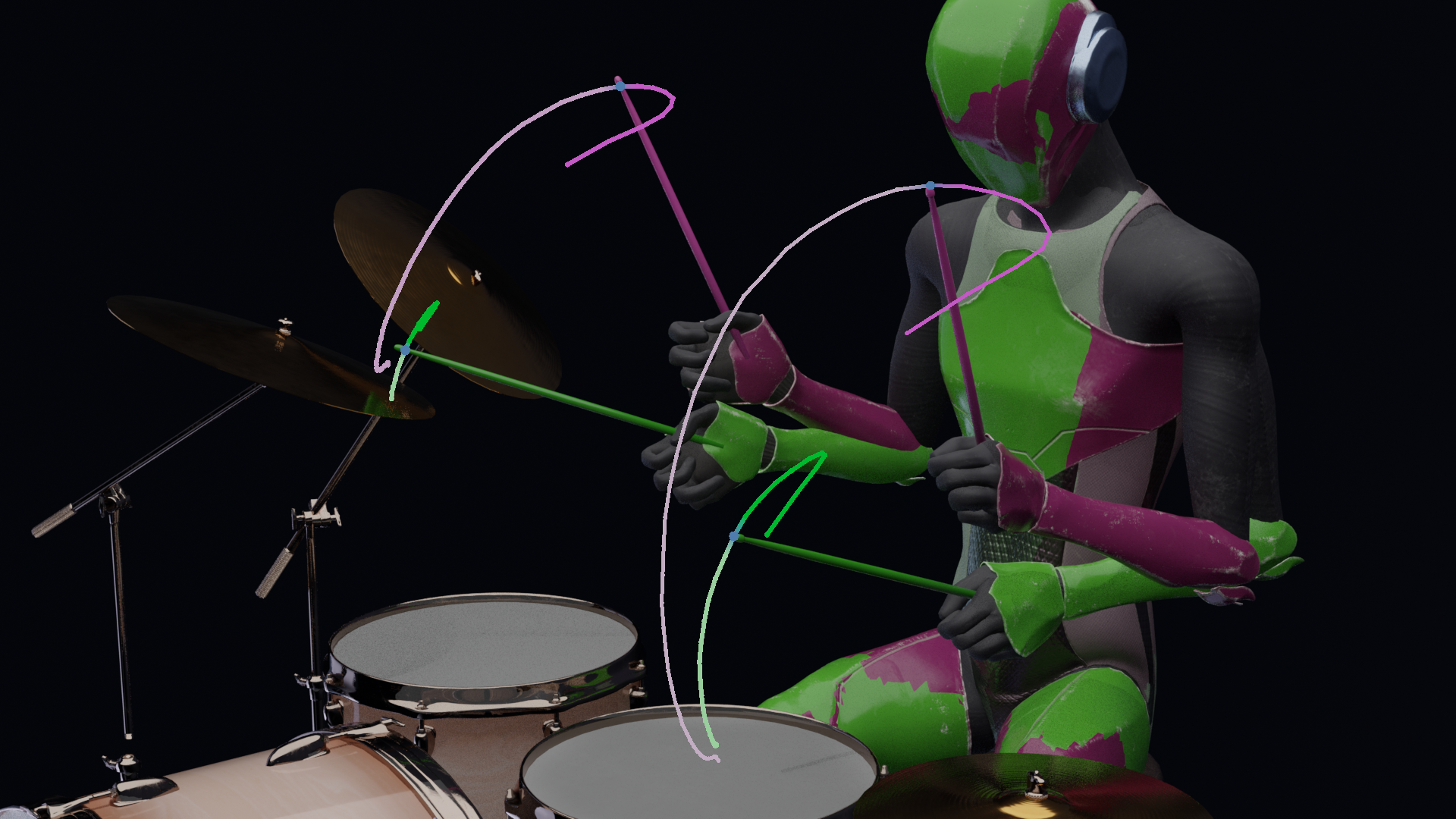}
  \caption{
    \textbf{Emergent gain modulation.}
    Motion generated by our dual-objective model with input audio at a normal volume (0 dB, pink) and a low volume (-15 dB, green).
    Loud audio produces with larger arcs, while softer audio produces gentle trajectories with smaller arcs.
    This behavior is learnt from our dataset's intensity diversity.
  }
  \label{fig.gain}
\end{figure}

\paragraph*{Gain modulation.}
Our dataset includes takes with similar content but different tempos and energy levels.
As a consequence, our model learned to adjust the amplitude of the motion based on the gain of the input audio: if the audio is louder, the performer will hit the drums harder.
We show an example of this intensity-control feature in Figure~\ref{fig.gain}.

\subsection{Application to Music Analysis: Motion-to-MIDI}\label{subsec.transcriber}
Beyond visual synthesis, the high fidelity of our generated motion allows for downstream tasks such as automatic drum transcription.
Music transcription is the task of converting music into a human-readable format, such as music scores or MIDI files.
Transcription is a critical tool in music education, analysis, and production -- for example, transcribing an instrumental track allows for seamless fine-grained editing.
However, automatic drum transcription remains particularly challenging due to the noisy and ambiguous nature of drum audio.
Existing polyphonic transcribers often struggle with drums, frequently misinterpreting percussive transients as noise~\cite{Wu2021,Gardner2021,Bittner2022,Maman2022}.
Moreover, state-of-the-art algorithms that specialize in drums are often limited to a narrow subset of drum components~\cite{Callender2020,Zehren2023}, typically focusing on snare, hi-hat, and kick.

Here, we introduce a proof-of-concept pipeline: \emph{motion-to-MIDI}, an unconventional yet highly effective approach to transcription.
By utilizing a two-step framework (audio-to-motion followed by motion-to-MIDI), our method first distills audio signals into cleaner continuous motion signals before identifying discrete events that form the MIDI signal.
Initial testing suggests that our motion-to-MIDI algorithm is able to transcribe drum tracks with higher fidelity than existing methods.
In qualitative comparisons, polyphonic tools such as Omnizart~\cite{Wu2021} and Spotify’s Basic-Pitch~\cite{Bittner2022} struggled significantly with percussive transients, often failing to produce coherent transcriptions.
While specialized models like ADTOF~\cite{Zehren2023} show stronger performance, our preliminary tests indicate that motion-to-MIDI more faithfully reproduces the ground truth, particularly when different drum components are involved.
Technical details and preliminary results are provided in Appendix~\ref{app.transcriber}.

While initial results on our internal dataset are promising, ongoing work includes extensive quantitative evaluation for rigorous comparison against existing state-of-the-art transcribers.


\section{Discussion}

\paragraph*{Comparison to prior work.}
Direct quantitative comparison to previous work~\cite{Kyriakou2025,Shahid2025} remains non-trivial due to discrepancies in datasets, modalities (audio vs MIDI), and evaluation methods. At the time of writing, public codebases or standardized precision metrics were unavailable for these works.
However, our approach offers three distinct advantages: (1) direct audio-driven synthesis, enabling real-world deployment without the need for transcription; (2) a fully generative architecture that avoids the stylistic and transitions constraints of motion-matching; and (3) the introduction of standardized metrics to facilitate systematic benchmarking in the field.

\paragraph*{Limitations.}
The primary limitation of our approach is the reliance on isolated drumming audio as input. For polyphonic music, the drum track must be extracted using source separation tools, such as Demucs~\cite{Defossez2019} or Spleeter~\cite{Hennequin2020}.
While the generated motion quality is tied to the fidelity of the stem extraction, our experiments suggest that current splitters are sufficiently robust to provide high-quality inputs, provided the source contains clear percussive events.
We also note that our model assumes a fixed spatial drum configuration. Generalizing to different configurations (e.g., via additional data capture or geometric conditioning) is left as future work.

\paragraph*{PAS applicability beyond drumming.}
The core principle of PAS (soft correspondence between discrete audio events and acceleration peaks) is applicable to other percussive tasks requiring tight audio-motion synchrony, such as typing, clapping, or rhythmic footstepping.

\paragraph*{Future directions.}
Three promising avenues emerge from our work towards further elevating the expressivity and realism of digital performances.
First, increasing visual fidelity through finger-level motion and facial dynamics would significantly improve performer realism.
Note that finger-stick dynamics involve complex sliding and multi-point pivoting, making it challenging to infer stick-tip positions directly from finger motion.
Therefore, we believe that explicit stick-tip supervision would remain important for high-precision drumming even if high-quality finger motion were available.
Second, incorporating stylistic nuance, such as high-intensity punk energy or a laid-back jazz feel, via style transfer or expanded datasets would allow for more diverse and expressive musical expressions.
Third, developing manual authoring tools for arm-assignment overrides and temporal offsets would provide creators with the fine-grained control necessary for professional-grade post-production.
Combined, these advancements would bridge the gap between automated generation and intentional creative expression.
From an architectural standpoint, future work could explore two-stage models. While our single-stage model successfully generates high-quality drumming motion and generalizes well, similar to recent single-stage models for cello synthesis~\cite{Qiu2025}, two-stage architectures have recently improved motion quality in human-object interactions (see, e.g., Ron et al.~\cite{Ron2025}). It would be interesting to investigate whether a two-stage approach could further refine the generated motion by dividing the task: for example, first predicting stick-tip positions from audio and subsequently synthesizing the final motion based on these targets.


\section{Conclusion}

In this paper, we presented a scalable approach for synthesizing high-fidelity drumming motion directly from audio.
By employing a dual-objective optimization strategy, our method achieves exceptional temporal and spatial precision.
Furthermore, we introduced novel performance metrics for systematic benchmarking of percussive motion, which we used to validate our results and will serve as benchmarks for future works.
Additionally, our user study reveals that our generated motions are perceptually indistinguishable from ground truth, underscoring the realism of the generated performance.
A quantitative analysis also demonstrates that our approach maintains robust generalization across diverse audio sources and genres.
Beyond drumming, this work establishes a robust foundation for high-precision multimodal synthesis and provides a scalable tool for the next generation of automated virtual performances.


\section{Acknowledgments}

We thank Dominik Borer and Jakob Buhmann for discussions and technical support, and Violaine Fayolle and Doriano van Essen for artistic support throughout the project. We also thank our professional drummer Brian Quinn (qtrio.ch).


\balance
\printbibliography            

\newpage
\onecolumn

\appendix

\section{Audio Features: Engineered vs Learned}\label{app.audio-feats}
Our system takes an drum audio track as input. The first step is to extract audio features that will serve as condition for the diffusion process.
Some multimodal models~\cite{Tseng2023} use a pre-trained feature extractor, such as Jukebox~\cite{Dhariwal2020}.
This approach is common for models that deal with complex audio, including voices and polyphonic music.
Since we focus on drums audio, we take a more traditional approach by separating the drum stem from the audio and later extracting human-engineered features.
This design choice leads to more explainable data preprocessing and a more maintainable pipeline that does not depend on other codebases or models being up-to-date.
While the choice of features may be further optimized, we found that our 44 chosen features are sufficient to generate high-quality motion, as shown in the main text and in the Supplementary Video.

\section{Calculation of Impact Point Deviations}\label{app.IPD}
The primary challenge in extracting impact points from drumming motion data is the lack of hand-specific labels (left vs right hand) at each audio event. While MIDI events provide exact timestamps, they do not specify which stick triggered the hit. Consequently, at each MIDI event, the spatial coordinates of both sticks are recorded as \emph{candidate impact points}. To isolate the true contact points from the spurious data generated by the non-striking hand, we considered two approaches: (1) manual/procedural annotation of all MIDI events as left or right hits and (2) filtering of candidate impact points via a clustering algorithm.
Given the efficiency and efficacy of the latter, we implemented a density-based clustering approach.

Our density-based clustering algorithm works as follows.
For each drum component (e.g., snare, hi-hat), we identify the candidate point with the highest local density, defined as the point with the maximum number of neighbors within a fixed radius $r$.
This high-density coordinate is established as the cluster center. We then apply a spatial threshold, retaining only those points within a distance $d$ from the center and discarding the outliers corresponding to the non-striking hand.
Through empirical validation, we determined that a radius of $r=3$ cm effectively identifies the core impact zone of each drum component while remaining robust against sparse spatial noise.
The discarding threshold was set at $d=7$ cm, resulting in a maximum cluster diameter of 14 cm. 
This value was selected to satisfy two primary physical constraints: (1) the 14 cm span is narrower than the $\sim25$ cm drum diameter, preventing cluster overlap with adjacent instruments, and (2) this threshold accommodates the $\sim8.3$ cm maximum inter-frame displacement calculated from peak stick velocities ($\sim10$ m/s \cite{Dahl2004}) at a 120 fps.
These values provide a sufficient buffer to capture high-velocity impacts while effectively filtering out the non-striking hand.

\begin{figure}[th]
  \centering
  \includegraphics[width=0.85\linewidth]{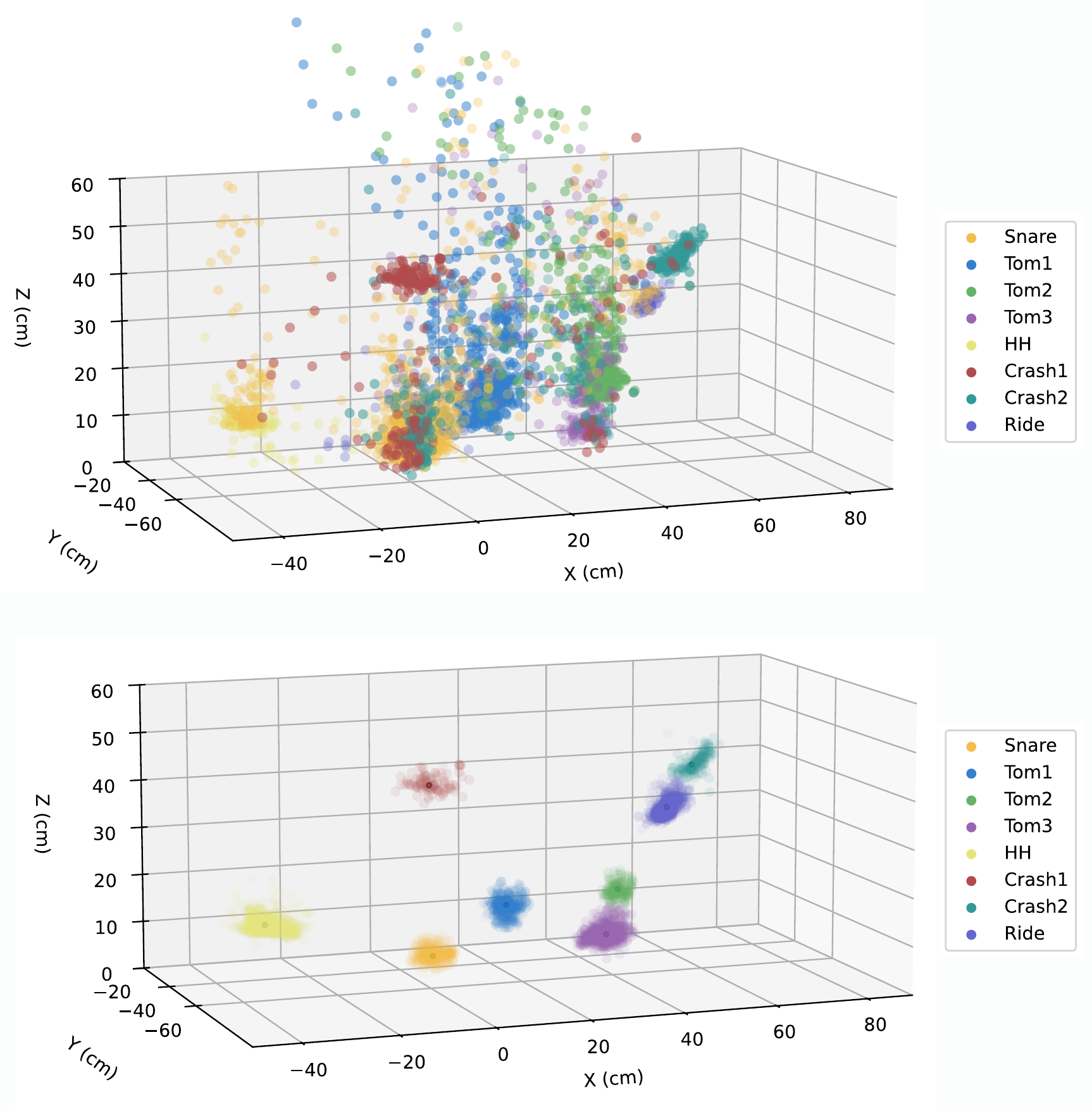}
  \caption{
    \textbf{Our clustering algorithm filters out noise to find the impact points for each drum component.}
    Candidate impact points (top) and filtered impact points (bottom).
    Cluster centroids indicated as a black marker.
    Data corresponding to our ground-truth test set.
  }
  \label{fig.clustering}
\end{figure}

Figure~\ref{fig.clustering} illustrates the robust performance of the algorithm in identifying the impact points for each drum component. In this example, we use the data from our test set.
The clusters exhibit vertical dispersion (rather than a perfect flat shape over the drum surface) due to temporal quantization at 120 fps, which frequently captures the stick position in a frame immediately preceding or following the physical impact.
Despite this variance, the morphology of each component remains apparent: snare, toms, and hi-hat appear as dense clusters with flat bottoms, while crashes and ride display diagonal orientations that reflect their physical mounting angles and striking surfaces.

We also evaluated an iterative centroid-based pruning method. This algorithm involves calculating the global centroid of all candidate points (for a given drum component), removing the most distant outliers, and recalculating the centroid until all remaining points fall within a predefined distance threshold.
However, this method proved suboptimal due to its high sensitivity to noise: since the set of candidate points contains a large number of ``wrong-hand'' points distributed far from the drum surface, the initial centroid was frequently pulled away from the true impact zone. This often led the algorithm to converge on a false center or fail to exclude relevant outliers, whereas the density-based approach remained robust against such spatial noise.

\begin{figure}[th]
  \centering
  \includegraphics[width=0.95\linewidth]{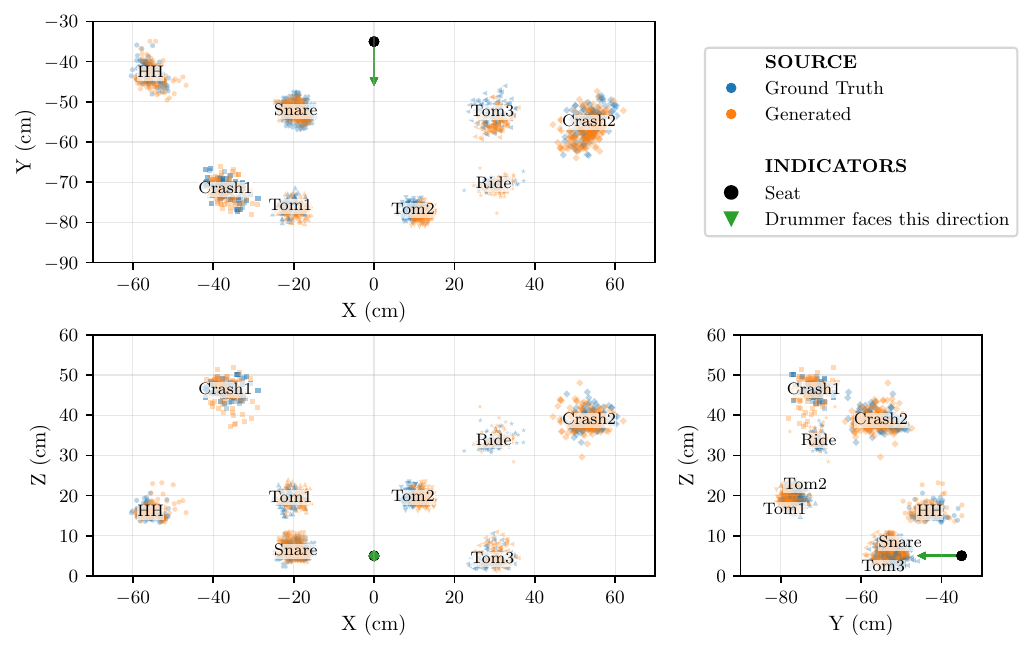}
  \caption{
    \textbf{Impact-point clusters of generated motion and ground truth overlap.}
    Clusters for the different drum components projected onto the XY (top-down), XZ (front-facing), and YZ (side-profile) planes. Blue points represent ground truth data, while orange points indicate generated impact distributions. The central black dot and green arrow indicate the drummer's seat position and facing orientation, respectively. All units are in centimeters (cm).
    The set of impact points shown here was chosen at random from our test set.
  }
  \label{fig.clusters}
\end{figure}

Figure \ref{fig.clusters} shows the impact-point clusters for ground truth (blue) and generated motion (orange) in our test dataset.
The overlap between blue and orange clusters suggests that the model accurately captures the size and shape of each instrument, reflecting the natural variability in a drummer's reach and striking motion during performance.

\section{Further details on the Percussive Alignment Score}\label{app.DAS}
Here, we discuss some design aspects of our temporal precision metric, the Percussive Alignment Score (PAS). First, we discuss how to detect audio and motion onsets.
Then, we explain how to choose the values of $\alpha$ and $\beta$.
Lastly, we discuss some basic properties of the PAS.

\subsection*{Detecting audio and motion onsets}\label{app.onsets}
\begin{figure}[th]
  \centering
  \includegraphics[width=0.8\linewidth]{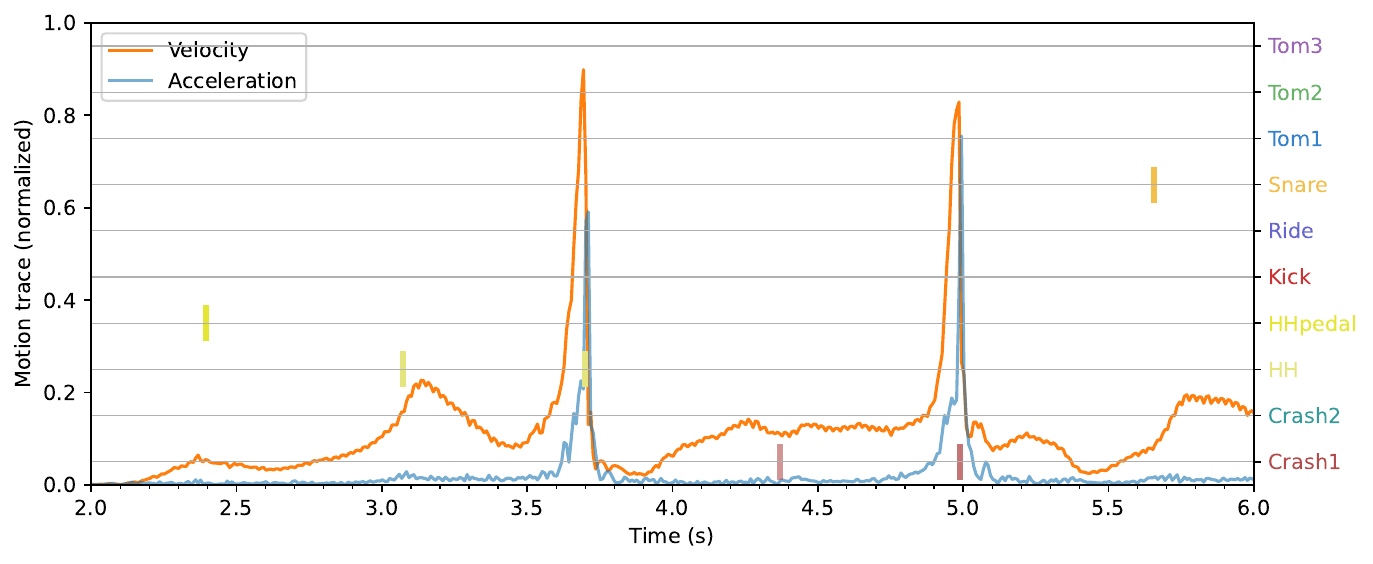}
  \caption{
    \textbf{Peaks in stick acceleration are more clearly correlated with audio events than peaks in velocity.}
    Magnitude of the velocity (orange) and the acceleration (blue) of the tip of the right stick, in one of the ground-truth motion sequences.
    Note activations from the MIDI signal are indicated as colored bars (left vertical axis).
    The right stick hits the hi-hat at $\sim 3.6$~s and the crash at $\sim 5$ s.
    The hi-hat pedal is activated with the foot, while the first hi-hat, the first crash, and the first snare hits are triggered with the left stick.
  }
  \label{fig.vel-acc-midi}
\end{figure}

Let us discuss audio and motion onsets separately:
\begin{itemize}
  \item \textbf{Audio onsets.}
  Audio onsets can be extracted from the audio or the MIDI signals. However, detecting onsets in an audio wave is nontrivial and prone to noisy detections.
  Hence, whenever a MIDI file is available (e.g., when working with our dataset), it is preferable to use the MIDI \emph{note-on} events as audio onsets.
  Additionally, each MIDI event also reports which drum component was triggered and a \emph{velocity} value in the range 0-127, which indicates how hard the component was hit.
  \item \textbf{Motion onsets.} The kinematics of the tips of the drumsticks share a strong correlation with the drums being hit. In particular, acceleration displays clear peaks on each impact. Therefore, we use the peaks in the acceleration of the stick tips as our motion onsets. 
  Figure~\ref{fig.vel-acc-midi} shows the velocity and the acceleration of the tip of a drumstick in one of the ground-truth sequences.
  This drumstick only impacts the drums twice: at 3.7 s and 5 s.
  The acceleration (blue) only shows clear peaks at these events.
  However, the velocity (orange) also shows peaks when the stick is not impacting a drum (e.g., at 3.2 s).
  Peaks can be computed using the \texttt{find\_peaks()} function from \texttt{scipy}.
\end{itemize}

\subsection*{Choosing the values of $\alpha$ and $\beta$}\label{app.alphabeta}
The PAS definition provided in the main text involves two parameters $\alpha$ and $\beta$.
Both parameters model the shape of a generalized Gaussian distribution applied to each term of the sum.
In our work, we use $\alpha = 40$ ms and $\beta=7$.

On the one hand, $\alpha$ controls the width of the Gaussian, i.e., how close the audio and motion onsets must be for the exponential terms to be close to 1.
If $\alpha$ is too small, it penalizes natural delays (the motion peak should always precede the audio event by a small margin), which is undesirable. If $\alpha$ is too large, all the exponential terms will be close to 1 and noisy motion will score high.
In practice, we found that the delay between motion onset and audio event is generally below 20 ms, and therefore we choose $\alpha=40$ ms, to ensure that these natural delays will not introduce any penalty.

On the other hand, $\beta$ controls the sharpness of the transition from 1 to 0 in the Gaussian.
We choose the value of $\beta$ such that a delay of 20 ms, which is approximately the maximum delay between ground-truth motion peaks and audio events, will result in a score of $0.99$ or more. That is, we enforce $e^{-(20/\alpha)^\beta} > 0.99$. For simplicity, we choose an integer value for $\beta$. This yields $\beta = 7$.

\subsection*{PAS for jittery motion}\label{app.rangeDAS}
As discussed in the main text, the PAS takes values between 0 and 1.
Note that we assume that the motion signal is well-behaved: if we consider an uncorrelated noise signal with many peaks, the PAS would be large as there would always be an onset close to an audio event. To prevent this situation, motion should always be validated as reasonable before employing the PAS.
We experimented with a penalty factor, such as
$$\frac
    {\min\left({|\mathcal{A}|,|\mathcal{M}|}\right)+1}
    {\max\left({|\mathcal{A}|,|\mathcal{M}|}\right)+1},$$
which would would take a small value if the number of audio and motion onsets are very different.
However, we found that high-quality data (including ground truth) sometimes exhibits motion peaks that do not necessarily correlate with audio (such as a limb following the beat when the style is very energetic), and therefore such a penalty term would penalize all data, effectively squashing the PAS from $[0,1]$ in theory to approximately $[0,0.6]$ in practice, difficulting the comparison between ground truth and generated motion.

\section{Motion-to-MIDI for automatic drum transcription}\label{app.transcriber}
This section provides additional technical details regarding the motion-to-MIDI algorithm and a qualitative comparison of its performance against existing state-of-the-art (SOTA) methods.

\paragraph*{The algorithm}
The transcription process converts generated motion data into symbolic MIDI events.
As a preprocessing step, we utilize the ground truth (GT) data to define the physical layout of the drum kit. For each drum component $i$, we calculate its spatial centroid $c_i$. These centroids serve as the target coordinates for determining which component was hit.
The algorithm processes the predicted motion of the sticks and pedals separately to generate MIDI onsets and labels:
\begin{itemize}
    \item \textbf{Stick events.}
    First, we calculate the acceleration of each stick and identify local peaks to determine potential onset times.
    To account for the fact that physical motion precedes the resulting audio, we analyze the stick's position at the frame immediately before the onset time.
    We then calculate the distance between the stick and each drum centroid $c_i$. If the closest component is within a predefined distance threshold $\delta_i$, a MIDI event is recorded for that specific drum.
    \item \textbf{Pedal events.}
    We monitor the position along the vertical axis.
    In our dataset, a peak in the vertical position typically occurs approximately 0.1s before a hit, corresponding to the foot lifting off the pedal prior to triggering it. Identifying these peaks allows us to trigger Kick or Hi-Hat pedal MIDI events.
\end{itemize}

As a proof-of-concept, our current implementation focuses on temporal accuracy; consequently, velocities of recorded MIDI events are set to a fixed value.
Future iterations will integrate stick-velocity mapping at the moment of impact to capture the full dynamic range of the performance.

\begin{figure}[th]
  \centering
  \includegraphics[width=0.8\linewidth]{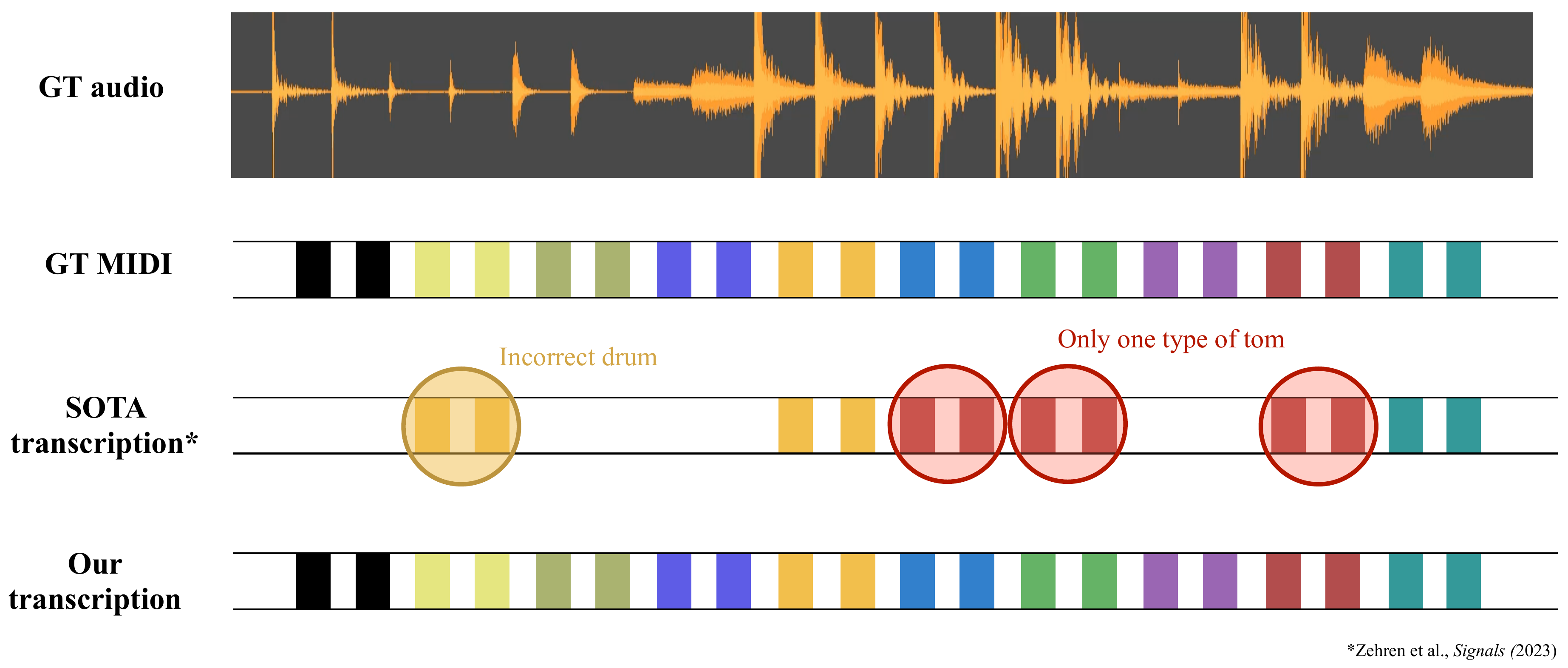}
  \caption{
    \textbf{Qualitative comparison of drum transcription performance}.
    The visualization displays the ground-truth (GT) audio waveform and the corresponding GT MIDI events. Each drum component is played twice, sequentially: kick (black), hi-hat pedal (light olive), hi-hat (dark olive), ride (blue purple), snare (yellow), tom 1 (blue), tom 2 (green), crash 1 (purple), floor tom (red), and crash 2 (aquamarine). The SOTA drum transcription~\cite{Zehren2023} exhibits some classification errors, such as identifying the incorrect drum and failing to differentiate between various toms. In contrast, our transcription, generated via the audio-to-motion and motion-to-MIDI pipeline, demonstrates high alignment with the ground truth across the entire instrument set.
  }
  \label{fig.midi}
\end{figure}

Figure~\ref{fig.midi} illustrates the transcription performance of our system on a representative sample.
The top two rows represent the ground truth waveform and its corresponding symbolic MIDI representation.
In this example, each drum component is triggered twice sequentially.
The third row displays the results from a state-of-the-art drums transcriber~\cite{Zehren2023}. In this example, some hits are not recorded while others are mapped to an incorrect drum.
The bottom row shows our algorithm's output. By leveraging motion as an intermediate representation, our model correctly identifies drum types and preserves the nuances of the performance, closely mirroring the GT MIDI.
While these qualitative results are promising, we are currently in the process of finalizing a comprehensive quantitative evaluation, extending beyond our test dataset, to ensure a fair and robust comparison between our method and SOTA transcribers.

\end{document}